\documentclass{article}
\usepackage[preprint]{corl_2026} 
\usepackage{graphicx,amsmath,amssymb}
\usepackage{xcolor}
\usepackage{array,colortbl}
\usepackage{adjustbox}
\usepackage{wrapfig}
\usepackage{booktabs}
\usepackage{multirow}
\usepackage{placeins}
\usepackage{makecell}

\newcommand{\systemname}{GEAR-VLA}

\title{GEAR-VLA: Learning Geometry-Aware Action Representations for Generalizable Robotic Manipulation}

\author{
{\normalfont\mdseries
\begin{tabular}{c}
\textbf{Yuan Zhang}$^{1,3*}$,
\textbf{Shiqi Zhang}$^{2*}$,
\textbf{Yedong Shen}$^{2}$,
\textbf{Shuai Dong}$^{2}$,
\textbf{Jiajun Deng}$^{2}$, \\
\textbf{Xin Zhang}$^{2}$,
\textbf{Yuxuan Gao}$^{2}$,
\textbf{Jiajia Wu}$^{3}$,
\textbf{Xin Nie}$^{3}$,
\textbf{Zhiyuan Cheng}$^{3}$,
\textbf{Jianmin Ji}$^{2}$, \\
\textbf{Yanyong Zhang}$^{2}$, \textbf{\textit{Fellow, IEEE}},
\textbf{Xingyi Zhang}$^{1\dagger}$, \textbf{\textit{Fellow, IEEE}},
\textbf{Jia Pan}$^{3\dagger}$ \\
[+0.3em]
{\small $^{1}$Anhui University \quad
$^{2}$University of Science and Technology of China \quad
$^{3}$iFLYTEK} \\ \\
[-0.3em]
\textbf{Project page:} https://babynabeauty.github.io/gear-vla-p/
\end{tabular}
}
}

\definecolor{oursgray}{gray}{0.93}

\begin{document}

\maketitle

\begingroup
\renewcommand{\thefootnote}{\fnsymbol{footnote}}
\footnotetext[1]{These authors contributed equally.}
\footnotetext[2]{Corresponding author.}
\endgroup

\begin{abstract}
Vision-Language-Action (VLA) models achieve strong benchmark performance but still struggle in real-world deployment with unseen objects, background shifts, and different robot embodiments. We argue that this stems from the lack of a unified geometry-aware manipulation representation, leaving existing VLAs vulnerable to low-level trajectory supervision, misaligned 3D features, and embodiment differences. To address this, we propose \systemname, a VLA framework for learning unified geometry-aware action representations for generalizable robotic manipulation. \systemname\ adopts coarse-to-fine action learning, where multi-source embodied pretraining equips the VLM with embodied reasoning and discrete action understanding before latent action tokens connect action semantics to a gradient-decoupled DiT continuous action expert. It further performs semantic-aligned 3D integration by aligning a trainable 3D spatial backbone with the VLA representation while freezing the original VLM-aligned visual pathway. To share this representation across robots, \systemname\ uses embodiment canonicalization, where embodiment-aware states and embodiment-invariant actions confine robot differences to the low-level interface. Extensive simulation and real-world experiments demonstrate strong generalization: \systemname\ achieves state-of-the-art performance on LIBERO, zero-shot LIBERO-Plus, and RoboTwin 2.0, reaches 85.9\% success on AgileX and 81.0\% on the pretraining-unseen LDT-01 embodiment, and obtains 90.1\% success on a 6,360-trial universal grasping benchmark with 212 unseen objects. 
\end{abstract}

\section{Introduction}
\label{sec:intro}

\begin{figure}[thbp]
  \centering
  \includegraphics[width=\linewidth]{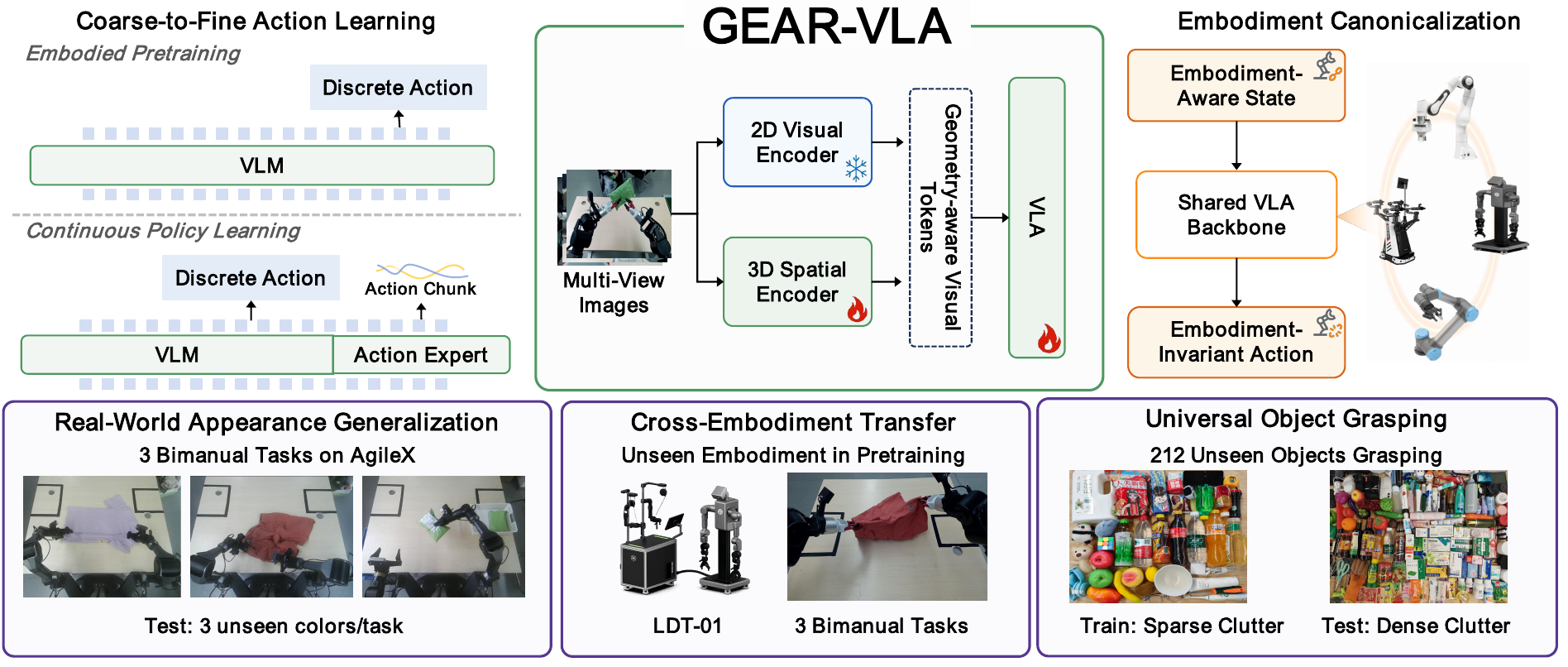}
  \caption{\small{GEAR-VLA learns geometry-aware action representations through three key designs: coarse-to-fine action learning, semantic-aligned 3D integration, and embodiment canonicalization. We evaluate these capabilities across real-world appearance generalization, cross-embodiment transfer, and universal object grasping.}}
  \label{fig:intro_overview}
  \vspace{-5mm}
\end{figure}

Vision-Language-Action (VLA)~\cite{openvla2025, pi02024, pi052025} models have emerged as a promising paradigm for building general-purpose robotic manipulation policies.
However, despite strong benchmark performance, current VLA systems still struggle in real-world deployment, where robots face unseen objects, background shifts, and different embodiments.

Existing efforts have advanced VLA generalization along three axes: action representation for making continuous control compatible with VLM-style token prediction, spatial representation for geometry-aware manipulation, and embodiment representation for scaling policies across heterogeneous robots. However, each axis still leaves a key representation gap. First, action-token-based VLAs~\cite{openvla2025,discrete_diffusion_vla2025,pi052025} tokenize continuous control, but because these tokens are largely quantized from low-level trajectories, they may push the VLM toward trajectory imitation rather than embodied reasoning. Second, VLAs need 3D spatial understanding, but existing depth~\cite{yuan2025depthvla}, 3D position encodings~\cite{qu2025spatialvla}, or spatial foundation priors~\cite{yang2026abot} are not naturally aligned with the VLM semantic space. As a result, they either remain action-head conditions or risk disturbing semantic representations when directly injected into the VLM backbone. Third, VLAs should transfer efficiently across robot embodiments. However, robot-specific action heads~\cite{bi2026hrdt} or embodiment prompts~\cite{xvla2025} can entangle robot identity with the shared policy representation, reducing transfer efficiency when robot data are imbalanced or the target embodiment is unseen.

To address these challenges, we propose \systemname, which learns unified geometry-aware action representations for generalizable manipulation. The central idea is to learn a unified action representation that is both semantically grounded and geometry-aware, while confining embodiment-specific adaptation to low-level interfaces.
Concretely, \systemname\ adopts a coarse-to-fine action learning strategy. During embodied VLM pretraining, FAST-style action tokens provide discrete supervision from robot trajectories, while latent action IDs extract high-level action semantics from manipulation videos. The resulting action-semantic representation is then connected through latent action tokens to a gradient-decoupled DiT-based action expert, allowing the model to produce continuous actions without back-propagating continuous action losses into the VLM backbone. To incorporate 3D geometry, \systemname{} aligns a trainable spatial backbone with the VLA representation. We freeze the VLM-aligned visual pathway and gradually inject 3D spatial features through a zero-initialized connector, enabling action prediction to leverage both 2D semantic cues and 3D geometric structure. To share this geometry-aware representation across robots, \systemname{} uses embodiment canonicalization, where embodiment-aware state inputs and an embodiment-invariant action space confine robot differences to the low-level interface instead of the shared VLA representation.

Extensive evaluations demonstrate strong generalization across simulation and real-world. \systemname{} achieves state-of-the-art performance on LIBERO~\cite{liu2021libero}, zero-shot LIBERO-Plus~\cite{fei25libero-plus}, and RoboTwin 2.0~\cite{chen2025robotwin}. In real-world experiments, it reaches 85.9\% average success across three contact-rich bimanual tasks on AgileX and 81.0\% success after adaptation to LDT-01, a robot embodiment with no similar counterpart in pretraining. We further construct a large-scale universal grasping benchmark with 212 unseen objects and three challenging real-world scenes, totaling 6,360 trials per method. \systemname{} achieves 90.1\% success, outperforming $\pi_{0.5}$~\cite{pi052025} (79.1\%) and DexGraspVLA~\cite{dexgraspvla2026} (84.4\%).

\begin{itemize}
   

    \item We introduce \systemname, a VLA framework that learns unified geometry-aware action representations through coarse-to-fine action learning and semantic-aligned 3D integration.

    \item We propose an embodiment-canonicalized state-action interface that transfers the learned representation across robot embodiments without robot-specific semantic prompts.

    \item Extensive evaluations on simulation benchmarks, real-world bimanual manipulation, cross-embodiment transfer, and universal grasping demonstrate strong generalization.

\end{itemize}

\section{Related Work}
\label{sec:related_work}

\textbf{Vision-Language-Action Models.}
Transformers~\citep{vaswani2017attention}, visual pretraining~\citep{dosovitskiy2021vit,radford2021clip}, and visual instruction tuning~\citep{llava2023} have enabled policies that map vision and language observations to robot actions.
Early systems combine language models with modular planning or affordance checking~\citep{asmussen2023saycan,liang2022codeaspolicies}, while RT-1, RT-2, PaLM-E, OpenVLA, and $\pi_0$ scale end-to-end VLA control with robot data and pretrained VLMs~\citep{brohan2022rt1,zitkovich2023rt2,driess2023palme,openvla2025,pi02024}.
Benchmarks such as Open X-Embodiment, LIBERO, and CALVIN~\citep{openx2024,liu2021libero,mees2021calvin} highlight the need for generalization across scenes, objects, tasks, and embodiments.

\textbf{3D Geometry for Embodied Perception.}
VLMs provide strong 2D semantic representations, but limited 3D spatial understanding for embodied interaction~\citep{3d_llm2023}.
Classical reconstruction systems such as COLMAP~\citep{schoenberger2016sfm,schoenberger2016mvs}, learning-based SfM methods such as VGGSfM~\citep{vggsfm2024}, monocular geometry models~\citep{depth_anything2024,moge2025}, and feed-forward 3D models~\citep{dust3r2024,mast3r2024,vggt2025} provide increasingly strong geometric cues.
Recent 3D-aware embodied models incorporate depth, point features, or spatial priors to improve robot grounding, but aligning such geometry with VLM token spaces remains challenging.

\textbf{Action Representation and Cross-Embodiment Control.}
Learning actions from heterogeneous data requires bridging low-level robot trajectories and broader visual dynamics from human or action-free videos.
Prior work studies latent action pretraining~\citep{latent_pretraining2025}, action tokenization such as FAST~\citep{fast2025}, and diffusion or flow-based policies for continuous control~\citep{chi2023diffusionpolicy,pi02024}.
Cross-embodiment learning further requires handling differences in kinematics, proprioceptive states, and action spaces; existing approaches often introduce embodiment-specific heads, adapters, or soft prompts~\citep{xvla2025}.

\section{Method}
\label{sec:method}
\begin{figure}[t]
  \centering
  \includegraphics[width=\linewidth]{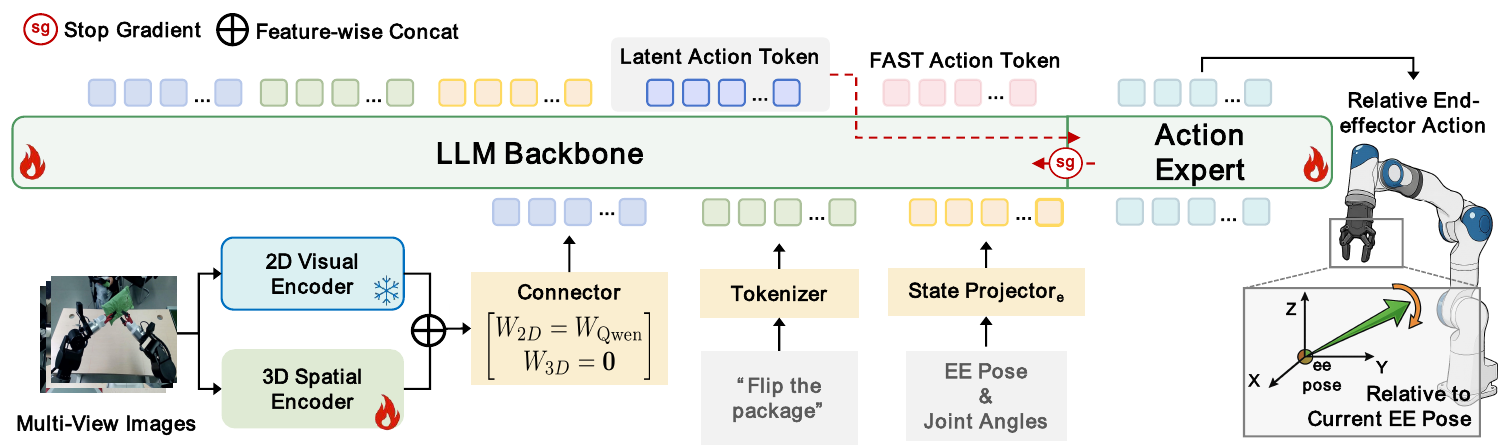}
  \caption{\small{Framework of \systemname{}.GEAR-VLA combines coarse-to-fine action learning, semantic-aligned 3D integration, and embodiment canonicalization to learn transferable geometry-aware action representations. Here, \(\mathrm{State\ Projector}_e\) denotes the lightweight state projector for robot embodiment \(e\).}}
  \label{fig:framework}
\end{figure}

\paragraph{Overview.}
\systemname{} is a VLA framework for learning unified geometry-aware action representations. Given multi-view RGB images $\mathcal{I}_t$, a language instruction $l$, and robot state $s_t$, the model predicts a future action chunk $\mathbf{A}_{t:t+H}$. As shown in Fig.~\ref{fig:framework}, GEAR-VLA consists of three core designs. Coarse-to-fine action learning first trains the VLM to acquire embodied reasoning and discrete action semantics, and then uses latent action tokens to condition a gradient-decoupled continuous action expert. Semantic-aligned 3D integration injects multi-view 3D structure while preserving the VLM semantic space. Finally, embodiment canonicalization uses embodiment-aware state and embodiment-invariant action space to confine robot-specific variation to the low-level interface. 

\subsection{Coarse-to-Fine Policy Learning}
\label{subsec:cea}
\textbf{Embodied VLM Pre-training.}
Directly training a VLA on low-level action trajectories can bias the model toward trajectory fitting. We therefore first perform embodied VLM pre-training, allowing the model to acquire embodied grounding, task understanding, and discrete action semantics before introducing the continuous action expert. We train the VLM on a mixture of embodied pre-training data, including vision-language understanding, spatial grounding, trajectory reasoning, pointing, mask tracking, and manipulation videos; the full data composition is provided in the appendix.

For action-labeled manipulation data, we use FAST-style action tokenization to convert continuous action sequences into FAST action tokens, which provide discrete supervision over robot action patterns. Inspired by VideoWorld2~\citep{ren2026videoworld2}, we further train a causal VQ-VAE latent action tokenizer to generate latent action IDs from manipulation videos.Unlike prior methods that learn discrete latents from single future-frame~\citep{wang2025vq,ye2025latent}, our tokenizer models continuous video segments and produces temporally continuous latent code sequences for short-horizon latent action dynamics. Since this requires only raw video frames, latent action IDs can be generated from robot, human, and web videos without action labels, and are used as VLM pre-training targets for learning action-relevant visual dynamics. We formulate all pre-training tasks as autoregressive token prediction, enabling the VLM to learn a shared representation of embodied semantics, spatial grounding, and discrete action patterns before continuous control is introduced:
\begin{equation}
    \mathcal{L}_{\mathrm{VLM}}=-\sum_i \log p_\theta(y_i \mid y_{<i}, O, l, s),
\end{equation}
where \(O\), \(l\), and \(s\) denote multi-view observations, language instruction, and robot state, respectively, and \(y_i\) denotes any target token, including text, grounding, planning, FAST action, or latent action.

\textbf{Continuous Action Generation. }
After the VLM has acquired embodied reasoning and discrete action semantics, we attach a DiT-based action expert to generate continuous action trajectories. The DiT does not access the full VLM representation; instead, it only uses the K/V cache of the latent action tokens produced by the VLM. Since these tokens have absorbed FAST-style robot action supervision and latent visual dynamics from action-free videos in the first stage, they serve as a compact semantic bridge between the VLM and the continuous action expert.

We train the DiT action expert with a flow matching loss and prevent this loss from updating the VLM backbone through gradient decoupling. Let \(A\) be the target continuous action chunk, \(\epsilon\) be Gaussian noise, \(\tau\in[0,1]\) be the flow time, and \(h_{\mathrm{la}}\) be the latent action token cache from the VLM. The objective is:
\begin{equation}
    \mathcal{L}_{\mathrm{FM}}
    =
    \mathbb{E}_{A,\epsilon,\tau}
    \left[
    \left\|
    v_\phi(A_\tau, \tau, \mathrm{sg}(h_{\mathrm{la}}))
    -
    (A-\epsilon)
    \right\|_2^2
    \right],
    \quad
    A_\tau = \tau A + (1-\tau)\epsilon .
\end{equation}
Here, \(\mathrm{sg}(\cdot)\) denotes stop-gradient. This prevents continuous control optimization from disrupting the visual-language understanding, spatial grounding, and discrete action semantics learned by the VLM. Thus, continuous action generation is not a simple regression head attached to the VLM; instead, the latent action token cache forms a controlled interface that stably maps discrete action semantics to executable continuous control.

\subsection{Semantic-Aligned 3D Geometry Integration}
\label{subsec:geom}
To introduce 3D geometric awareness while preserving the semantic understanding of the VLM, we add a trainable 3D spatial encoder alongside the original 2D visual pathway. Since the original 2D visual encoder has been aligned with the LLM representation space through large-scale vision-language pretraining, we freeze this branch to preserve stable language-aligned visual representations. We use VGGT as the 3D spatial encoder. Compared with single-frame depth or explicit 3D position encodings, VGGT exploits multi-view consistency to model scene layout, object shape, and spatial relations, providing geometric structural cues useful for manipulation. However, VGGT features are not naturally adapted to the VLM semantic space, and directly injecting them may perturb the original visual-token distribution. Therefore, during the coarse-to-fine training process, we jointly train the 3D spatial encoder and the LLM backbone with large-scale spatial perception and embodied reasoning data.

Specifically, we concatenate the 2D visual features and 3D structural features along the feature dimension, and map them into the LLM visual-token space through an expanded visual projector:
\begin{equation}
    Z^{\mathrm{vis}}
    =
    [H^{2D}; H^{3D}] W_{\mathrm{vis}} + b,
    \quad
    W_{\mathrm{vis}}^{(0)} = [W_{\mathrm{Qwen}}; 0],
\end{equation}
where \(H^{2D}\) and \(H^{3D}\) denote the features extracted by the 2D visual encoder and the 3D spatial encoder, respectively, and \(W_{\mathrm{vis}}\) denotes the expanded visual projector. The 2D block is initialized from the original Qwen2.5-VL projector, while the newly added 3D block is zero-initialized, so that the model preserves the original VLM visual-token distribution at the beginning of training. As training proceeds, the 3D spatial encoder gradually learns manipulation-relevant geometric structure, enabling the unified representation space to combine 2D semantic cues with 3D geometric cues and form geometry-aware action representations.


\subsection{Embodiment Canonicalization}
\label{subsec:embcanon}
The previous two sections build a shared VLA representation with action semantics and geometric awareness through coarse-to-fine learning and 3D integration. If robot-specific prompts or separate action heads are used to distinguish different embodiments, robot identity may be injected into the high-level policy representation, weakening representation sharing and imposing stronger requirements on balanced data across robots. We therefore adopt Embodiment Canonicalization during training, enabling data from different robots to be learned and transferred within the same shared representation. On the action-output side, we use Relative End-effector Action, allowing different robots to share a unified relative end-effector action space. Let \(T_t^{ee}\in SE(3)\) denote the current end-effector pose, and \(T_{t+i}^{ee}\in SE(3)\) denote the \(i\)-th future target end-effector pose, both expressed in the robot base coordinate. We represent each future action as an SE(3) relative transform with respect to the current end-effector pose:
\begin{equation}
\Delta T_{t+i}
=
(T_t^{ee})^{-1} T_{t+i}^{ee},
\quad i=1,\ldots,K .
\end{equation}
The predicted action chunk is therefore
\begin{equation}
A_t^{rel}
=
\left[
\Delta T_{t+1}, \Delta T_{t+2}, \ldots, \Delta T_{t+K}, g_{t+1:t+K}
\right],
\end{equation}
where \(g_{t+1:t+K}\) denotes the gripper command sequence. All future poses are expressed relative to the same current end-effector pose \(T_t^{ee}\), rather than step-wise relative to the previous pose. Compared with step-wise delta actions, Relative End-effector Action avoids accumulating errors within an action chunk; compared with absolute actions, it reduces dependence on robot base coordinates and workspace-specific geometry.

On the state-input side, we use the end-effector pose and joint angles as paired state inputs: $s_t = \{T_t^{ee}, q_t\},$
where \(q_t\) denotes the joint angles. The correspondence between end-effector pose and joint angles implicitly captures embodiment information such as degrees of freedom, joint layout, and kinematic constraints. For each robot embodiment, we use only a lightweight state projector to map its state into the shared VLA representation space:
$z_t^{s} = f_{\psi_e}^{s}(s_t),$
where \(e\) denotes the robot embodiment and \(f_{\psi_e}^{s}\) is its corresponding state projector. When transferring to a new robot, we use two-stage lightweight adaptation: first aligning a new state projector with the shared VLA backbone frozen, and then performing light end-to-end fine-tuning. This brings the new robot's state distribution into the existing shared representation with minimal disturbance to the learned manipulation semantics.

\section{Experimental Results}
\label{sec:experimental_results}

\textbf{Dataset.}
We evaluate \systemname{} across simulation benchmarks and real-world robot settings. For simulation, we use LIBERO~\cite{liu2021libero}, LIBERO-Plus~\cite{fei25libero-plus}, and RoboTwin 2.0~\cite{chen2025robotwin}, which cover standard manipulation, structured out-of-distribution generalization, and large-scale multi-task simulation, respectively. For real-world evaluation, we consider bimanual manipulation on two robot embodiments and a large-scale universal grasping benchmark with unseen objects and cluttered scenes.

\begin{table*}[!htbp]
\centering
\caption{\small{\textbf{Simulation benchmark summary.} We report averages over LIBERO, zero-shot LIBERO-Plus, and RoboTwin-2.0 clean/randomized settings. We evaluate all methods under the same evaluation protocol when reproduced, and otherwise report numbers from official sources.}}
\label{tab:simulation_results}
\scriptsize
\renewcommand{\arraystretch}{1.15}
\setlength{\tabcolsep}{4pt}
\begin{tabular}{l|cccccccc}
\toprule
\textbf{Benchmark}
& \textbf{OpenVLA}~\citep{openvla2025}
& \textbf{WorldVLA}~\citep{worldvla2025}
& $\boldsymbol{\pi_0}$~\citep{pi02024}
& \textbf{UniVLA}~\citep{univla2025}
& $\boldsymbol{\pi_{0.5}}$~\citep{pi052025}
& \textbf{X-VLA}~\citep{xvla2025}
& \textbf{ACoT}~\citep{acotvla2026}
& \textbf{Ours} \\
\midrule
LIBERO Avg. 
& 76.5 & 81.8 & 94.4 & 95.5 & 96.9 & 98.1 & 98.5 & \textbf{98.7} \\
LIBERO-Plus Avg. 
& 15.6 & 25.0 & 69.4 & 42.9 & 85.7 & 68.3 & 86.6 & \textbf{88.7} \\
RoboTwin-2.0 Clean Avg. 
& 38.3 & 42.5 & 48.4 & 45.8 & 82.7 & 72.9 & 80.1 & \textbf{91.1} \\
RoboTwin-2.0 Rand Avg. 
& 26.7 & 32.2 & 26.4 & 35.1 & 76.8 & 72.8 & 78.7 & \textbf{89.9} \\
\bottomrule
\end{tabular}
\vspace{-2mm}
\end{table*}


\begin{figure}[thbp]
  \centering
  \includegraphics[width=0.9\linewidth]{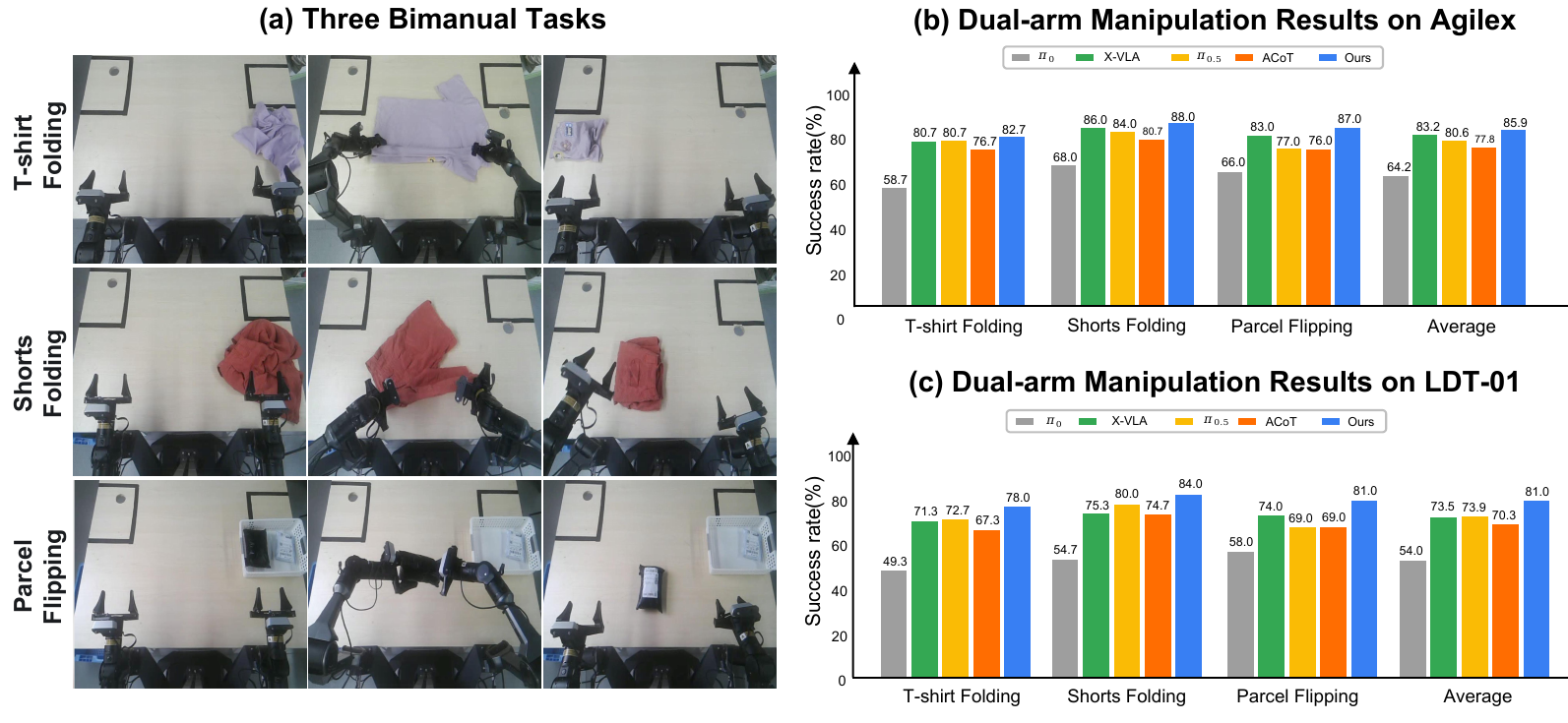}
  \vspace{-4mm}
  \caption{\small{Real-World Manipulation Results. (a) Three evaluation tasks: T-shirt folding, shorts folding, and parcel-label flipping. (b) Results on AgileX, where each task is trained with 200 demonstrations using one fixed object color and tested on three unseen colors. (c) Results on LDT-01, a pretraining-unseen 16-DoF bimanual robot. GEAR-VLA achieves the best performance on both platforms, showing strong real-world generalization.}}
  \vspace{-4mm}
  \label{fig:realWorldExp}
\end{figure}

\textbf{Implementation Details.}
All models are trained with bf16 mixed precision using AdamW and a base learning rate of $2\times10^{-5}$. Coarse-to-fine policy learning uses data-parallel training on 240 H200 GPUs, with per-GPU batch sizes of 8 and 4 for 350K and 700K iterations, respectively. Downstream finetuning is task-dependent; for LIBERO, we train on 56 H200 GPUs with per-GPU batch size 4 for 12K iterations. We use VGGT as the 3D spatial backbone, initialize the 2D visual encoder and LLM from Qwen2.5-VL, and freeze the semantic visual pathway. Robot trajectories are discretized with the FAST tokenizer, and latent action IDs are constructed with a causal VQ-VAE. All action-labeled data are resampled to 30\,Hz with a 30-step action chunk, corresponding to a 1-second prediction horizon. We use a constant learning-rate schedule with 3\% warmup, gradient checkpointing, and image augmentation. Additional details are in Appendix.


\subsection{Standard Simulation Benchmarks}
As shown in Table~\ref{tab:simulation_results}, GEAR-VLA achieves the best performance on all three benchmarks, showing strong in-distribution manipulation and robust generalization under simulation shifts. On LIBERO, it reaches 98.7\% average success, outperforming all prior methods. On LIBERO-Plus, it transfers zero-shot to out-of-distribution environments with seven perturbation settings, achieving 88.7\% success without adaptation and exceeding the strongest baseline by 2.1 points. On RoboTwin 2.0, GEAR-VLA achieves 91.06\% and 89.92\% success under clean and randomized settings, improving over ACoT by 11.00 and 11.20 points. These results suggest that GEAR-VLA learns a transferable manipulation representation with both action semantics and geometric structure, rather than fitting benchmark-specific trajectory patterns.

\subsection{Real-World Manipulation}
To validate GEAR-VLA in real-world settings, we design three challenging bimanual tasks as shown in Fig.~\ref{fig:realWorldExp} (a): T-shirt folding, shorts folding, and parcel-label flipping. 
Using the proposed embodiment canonicalization, we lightly adapt GEAR-VLA to two robot embodiments, AgileX and LDT-01. Notably, no embodiment similar to LDT-01 appears during pretraining, making this a realistic test of real-world transfer and cross-embodiment adaptation.

\begin{table*}[!htbp]
    \centering
    \vspace{-4mm}
    \caption{\small{\textbf{Real-world pick-and-place results across object categories and scene settings.} Avg. denotes the average over Sparse, Dense, and BG/Light settings. BG/Light denotes background and lighting variation.}}
    \label{tab:open_vocab_pick_place}
    \scriptsize
    \renewcommand{\arraystretch}{1.15}
    \setlength{\tabcolsep}{4pt}
    \begin{tabular}{lcccccccccccc}
    \toprule
    \multirow{2}{*}{\textbf{Object Category}} 
    & \multicolumn{4}{c}{$\boldsymbol{\pi_{0.5}}$~\citep{pi052025}} 
    & \multicolumn{4}{c}{\textbf{DexGraspVLA}~\citep{dexgraspvla2026}} 
    & \multicolumn{4}{c}{\textbf{Ours}} \\
    \cmidrule(lr){2-5} \cmidrule(lr){6-9} \cmidrule(lr){10-13}
    & \textbf{Sparse} & \textbf{Dense} & \textbf{BG/Light} & \textbf{Avg.}
    & \textbf{Sparse} & \textbf{Dense} & \textbf{BG/Light} & \textbf{Avg.}
    & \textbf{Sparse} & \textbf{Dense} & \textbf{BG/Light} & \textbf{Avg.} \\
    \midrule
    Axisymmetric & 86.3 & 80.1 & 79.2 & 81.9 & 90.8 & 85.4 & 83.1 & 86.4 & 92.4 & 90.2 & 89.5 & \textbf{90.7} \\
    Block-like             & 87.2 & 81.5 & 80.2 & 83.0 & 91.7 & 86.5 & 84.8 & 87.7 & 93.5 & 91.7 & 92.0 & \textbf{92.4} \\
    Irregular              & 76.5 & 67.7 & 65.6 & 69.9 & 82.1 & 75.1 & 74.2 & 77.1 & 88.4 & 86.7 & 84.9 & \textbf{86.7} \\
    Tool                   & 75.0 & 63.3 & 61.7 & 66.7 & 81.7 & 71.7 & 73.3 & 75.6 & 88.3 & 86.7 & 85.0 & \textbf{86.7} \\
    Bagged                 & 80.0 & 72.0 & 68.0 & 73.3 & 86.0 & 80.0 & 78.0 & 81.3 & 88.0 & 86.0 & 86.0 & \textbf{86.7} \\
    \midrule
    \textbf{Overall Avg.}  & 84.1 & 77.3 & 75.9 & 79.1 & 88.9 & 83.1 & 81.3 & 84.4 & 91.7 & 89.7 & 89.0 & \textbf{90.1} \\
    \bottomrule
    \end{tabular}
    \vspace{-2mm}
\end{table*}
\begin{figure}[thbp]
  \centering
  \includegraphics[width=0.9\linewidth]{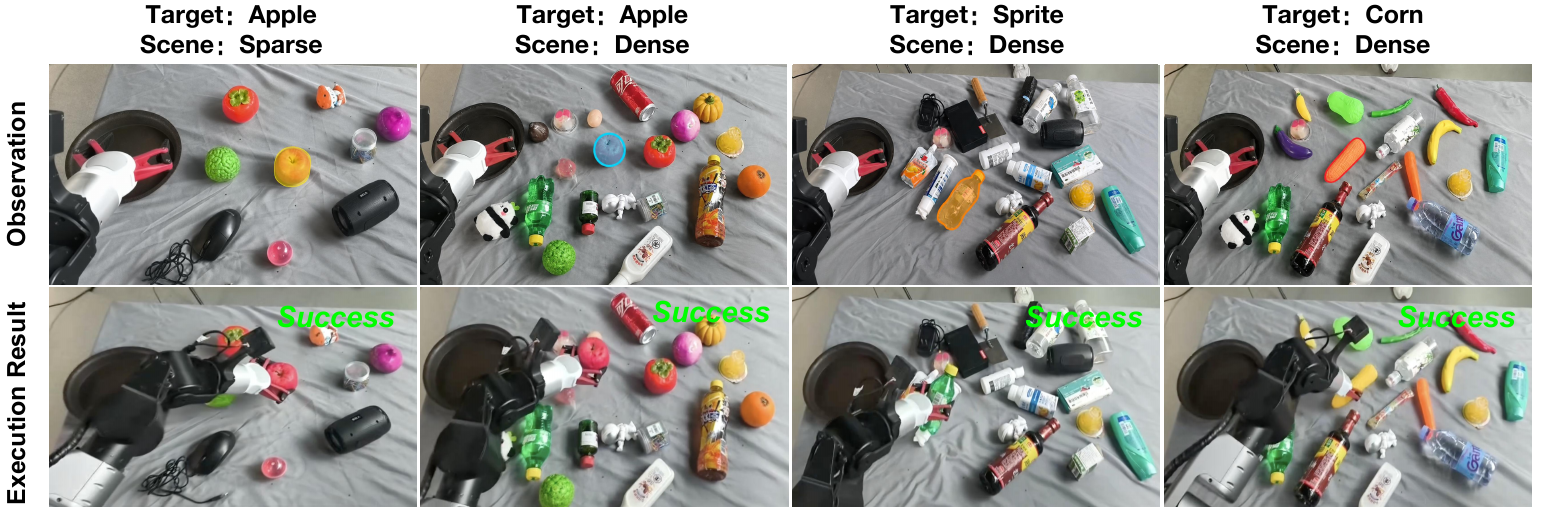}
  \label{fig:graspanything}
  \vspace{-2mm}
  \caption{\small{Qualitative results on the real-world universal grasping benchmark.} Each column shows the observation (top) and execution result (bottom). With only a first-frame target mask, GEAR-VLA localizes and manipulates diverse unseen objects in sparse and cluttered scenes.}
\end{figure}
\textbf{Real-World Generalization. }
We first evaluate GEAR-VLA on AgileX, a 14-DoF dual-arm robot. For the three tasks, we collect 200 bimanual teleoperation demonstrations per task using one fixed target-object color, yielding 600 training samples. To test zero-shot object variation, each task is evaluated on three unseen target-object appearances and sizes, requiring the model to rely on task semantics, object geometry, and spatial state rather than memorized color, texture, or scale cues. As shown in Fig.~\ref{fig:realWorldExp} (b), GEAR-VLA achieves 85.9\% average success across the three real-world tasks, outperforming all baselines and showing robust execution from semantic and geometric cues.


\textbf{Cross-Embodiment Transfer. }
To evaluate transfer to robot embodiments unseen during pretraining, we adapt GEAR-VLA to LDT-01, a 16-DoF bimanual robot with no similar counterpart in the pretraining data, making it a stricter test of cross-embodiment generalization. We collect only 200 demonstrations per task for lightweight adaptation under the same setup as AgileX, and evaluate T-shirt folding, shorts folding, and parcel-label flipping with unseen target-object appearances and sizes. As shown in Fig.~\ref{fig:realWorldExp} (c), GEAR-VLA outperforms all strong baselines on LDT-01, improving over $\pi_{0.5}$ by 7.1 points. This shows that the embodiment-canonicalized state-action interface confines embodiment-specific variation to low-level interfaces without disrupting the shared manipulation representation, enabling transfer to previously unseen robot embodiments.


\subsection{Large-Scale Universal Grasping}
To further evaluate GEAR-VLA under complex real-world conditions, we construct a large-scale universal grasping benchmark. 
\begin{table*}[!htbp]
\centering
\caption{\small{{\textbf{Ablation studies on LIBERO-Plus.} All entries are success rates (\%). The Discrete Action Learning group removes either latent action IDs from action-free videos or FAST tokens from robot trajectories. The 3D Geometry Integration group ablates VGGT, trainable VGGT adaptation, zero-initialized 3D projection, and frozen 2D ViT. The Embodiment Canonicalization group evaluates the two-stage adaptation strategy, embodiment-specific state projector, and X-VLA-style robot-specific soft prompts~\cite{xvla2025}.}}}
\label{tab:ablation_summary}

\scriptsize
\renewcommand{\arraystretch}{1.08}
\setlength{\tabcolsep}{1.45pt}

\begin{tabular}{
>{\centering\arraybackslash}m{0.125\textwidth}
>{\centering\arraybackslash}m{0.300\textwidth}
cccccccc
}
\toprule
\textbf{Group} & \textbf{Variant}
& \textbf{Cam.} & \textbf{Rob.} & \textbf{Lang.} & \textbf{Light}
& \textbf{Bg.} & \textbf{Noise} & \textbf{Layout} & \textbf{Avg.} \\
\midrule

\textbf{Ours}
& Full model
& \textbf{82.6} & \textbf{84.1} & \textbf{82.4} & \textbf{97.9}
& \textbf{93.1} & \textbf{90.0} & \textbf{89.4} & \textbf{88.7} \\
\midrule

\multirow{2}{*}{\makecell[c]{\textbf{Discrete Action}\\\textbf{Learning}}}
& w/o Latent Action IDs
& 81.2 & 82.4 & 79.7 & 95.8 & 91.8 & 87.9 & 88.9 & 87.1 {\color{red}{(-1.6)}} \\
& w/o FAST Tokens
& 79.7 & 80.6 & 78.4 & 95.2 & 91.2 & 89.1 & 87.8 & 86.2 {\color{red}{(-2.5)}} \\
\midrule

\multirow{4}{*}{\makecell[c]{\textbf{3D Geometry}\\\textbf{Integration}}}
& w/o VGGT
& 80.2 & 79.4 & 79.1 & 93.8 & 87.6 & 88.4 & 86.7 & 85.1 {\color{red}{(-3.6)}} \\
& Frozen VGGT
& 79.5 & 80.3 & 78.4 & 93.2 & 88.6 & 89.4 & 86.2 & 85.2 {\color{red}{(-3.5)}}  \\
& w/o Zero-Init 3D Projector
& 74.3 & 78.4 & 74.9 & 92.6 & 88.0 & 80.3 & 82.5 & 81.9 {\color{red}{(-6.8)}} \\
& Trainable 2D ViT
& 81.4 & 81.8 & 79.9 & 95.7 & 89.9 & 88.1 & 88.3 & 86.6 {\color{red}{(-2.1)}} \\
\midrule

\multirow{3}{*}{\makecell[c]{\textbf{Embodiment}\\\textbf{Canonicalization}}}
& One-stage adaptation
& 72.2 & 75.3 & 74.6 & 86.0 & 82.1 & 80.1 & 82.0 & 79.0 {\color{red}{(-9.7)}} \\
& w/ X-VLA Soft Prompt
& 78.2 & 79.4 & 78.9 & 93.9 & 90.4 & 86.9 & 86.1 & 85.0 {\color{red}{(-3.7)}} \\
& w/o Embodiment-Specific Projector
& 80.4 & 82.4 & 79.9 & 95.4 & 91.3 & 88.8 & 87.5 & 86.7 {\color{red}{(-2.0)}} \\
\bottomrule
\end{tabular}

\vspace{-6mm}
\end{table*}
The benchmark uses a referring mask-guided prompt to specify arbitrary objects and tests whether the model can correctly manipulate the object based on its spatial location and geometry. Since the VLM is pretrained with mask-tracking task, GEAR-VLA can use only the first-frame mask as a prompt and infer the target object in subsequent observations. The training set contains only 35 objects with 100 demonstrations each, collected with visually or geometrically similar distractors in the background. The test set contains 212 unseen objects across three settings: sparse clutter, dense clutter, and background/lighting variation. In each trial, we randomize the target position, orientation, and the number, category, and placement of background objects. Each object is tested 10 times, resulting in 2,120 trials per method per setting and 6,360 real-robot trials in total. Qualitative results are shown in Fig.~\ref{fig:graspanything}. As shown in Table~\ref{tab:open_vocab_pick_place}, GEAR-VLA achieves 90.1\% success, outperforming $\pi_{0.5}$ by 11.0 points and DexGraspVLA by 5.7 points. Unlike DexGraspVLA, which requires persistent target-mask tracking, GEAR-VLA uses only the first-frame target mask. Its large gains on irregular objects, tool objects, and dense scenes suggest stronger geometry-aware target grounding under heavy distractors.

\subsection{Ablation Studies}

We conduct ablations on LIBERO-Plus to evaluate each key design in \systemname, with results in Table~\ref{tab:ablation_summary}. In Discrete Action Learning, removing Latent Action IDs or FAST Token supervision decreases average success to 87.1\% and 86.2\%, respectively, showing the benefit of learning discrete action semantics from both action-free videos and robot trajectories. In 3D Geometry Integration, removing the 3D spatial encoder (VGGT) drops performance by 3.6 points, confirming the effectiveness of 3D structural features. Freezing VGGT also reduces average success to 85.2\%, suggesting that fixed 3D features are insufficient and the 3D spatial encoder must adapt to the VLA representation space. Randomly initializing the newly added 3D projector causes a larger 6.8-point drop, indicating that injecting unaligned 3D features can destabilize learning. Making the original 2D ViT trainable further underperforms the full model, showing that preserving the VLM-aligned 2D visual pathway is important for stable semantic-geometric fusion. In Embodiment Canonicalization, one-stage adaptation to an unseen robot reduces success to 79.0\%, indicating a mismatch between the new robot's state distribution and the shared VLA representation. Removing the embodiment-specific state projector leads to a 2.0\% drop, while adding X-VLA-style soft prompts~\citep{xvla2025} decreases performance by 3.7\%, suggesting that low-level state adaptation is more effective than injecting robot-specific semantic prompts into the shared representation.
\section{Conclusion}
\label{sec:conclusion}


We presented \systemname, a VLA framework for learning unified geometry-aware action representations for generalizable robotic manipulation. By combining coarse-to-fine action learning, semantic-aligned 3D integration, and embodiment canonicalization, \systemname\ enables VLMs to acquire action semantics, leverage 3D geometric structure, and transfer across robot embodiments through low-level state-action interfaces. Experiments across simulation, real-world bimanual manipulation, cross-embodiment transfer, and universal grasping demonstrate strong generalization to unseen objects, scenes, and robot embodiments.






\bibliography{example}  
\clearpage
\appendix
\newcommand{\appendixsection}[1]{%
\refstepcounter{section}%
\section*{Appendix~\thesection\quad #1}%
}

\appendixsection{Training Data and Implementation Details}
\label{sec:appendix}
\label{sec:appendix_training}

\paragraph{Training data.}
Tables~\ref{tab:general_embodied_perception_data} and
\ref{tab:robotic_human_manipulation_data} summarize the two data pools used in
Coarse-to-Fine Policy Learning. Table~\ref{tab:general_embodied_perception_data}
reports general vision-language and embodied perception data, including
vision-language understanding, spatial grounding, trajectory reasoning,
pointing, affordance understanding, and mask tracking.
Table~\ref{tab:robotic_human_manipulation_data} reports robotic and human
manipulation data, including action-labeled robot trajectories and action-free
human manipulation videos.

Embodied Pretraining uses both data pools: the full general vision-language and
embodied perception data in Table~\ref{tab:general_embodied_perception_data},
together with the robotic and human manipulation data in
Table~\ref{tab:robotic_human_manipulation_data}. All objectives are formulated
as autoregressive token prediction. For action-labeled robot trajectories, we
use both latent action IDs and FAST-style action tokens, so the VLM learns
high-level action semantics and discrete robot action patterns. For action-free
human manipulation videos, only latent action IDs are used, allowing the model
to learn manipulation dynamics without robot action labels.

Continuous Policy Learning uses the full robotic and human manipulation data in
Table~\ref{tab:robotic_human_manipulation_data}, together with 40\% of the
general vision-language and embodied perception data in
Table~\ref{tab:general_embodied_perception_data}. This preserves
visual-language understanding, spatial grounding, and discrete action semantics
while connecting the learned action representation to the gradient-decoupled
DiT-based action expert. The VLM continues to receive autoregressive supervision
from latent action IDs and FAST-style action tokens, while the DiT action expert
is trained with a flow matching loss under gradient decoupling. Quantities in
Table~\ref{tab:general_embodied_perception_data} are reported in thousands of
samples, while quantities in Table~\ref{tab:robotic_human_manipulation_data}
are reported in hours. In both tables, self-built denotes datasets constructed
by us for this work.

\begin{table}[!htbp]
\centering
\caption{\textbf{General vision-language and embodied perception data.} Quantities are reported in thousands of samples. Open-source denotes publicly available data, and self-built denotes datasets constructed by us for this work.}
\label{tab:general_embodied_perception_data}
\scriptsize
\setlength{\tabcolsep}{6pt}
\renewcommand{\arraystretch}{0.98}
\resizebox{0.58\textwidth}{!}{
\begin{tabular}{lcc}
\hline
\textbf{Type} & \textbf{Quantity} & \textbf{Source} \\
\hline
General VQA & 132k & Open-source \\
Understanding QA & 578k & Open-source + self-built \\
Planning QA & 35k & Open-source + self-built \\
2D Trajectory & 698k & Self-built \\
3D Trajectory & 150k & Self-built \\
2D Grounding & 100k & Open-source \\
3D Grounding & 632k & Self-built \\
Space Pointing & 625k & Self-built \\
Object Pointing & 401k & Self-built \\
Affordance & 393k & Self-built \\
Mask Tracking & 445k & Self-built \\
\hline
\end{tabular}
}
\end{table}

\begin{table}[!htbp]
\vspace{1mm}
\centering
\caption{\textbf{Robotic and human manipulation data.} Quantities are reported in hours. Open-source denotes publicly available data, and self-built denotes datasets constructed by us for this work.}
\label{tab:robotic_human_manipulation_data}
\scriptsize
\setlength{\tabcolsep}{6pt}
\renewcommand{\arraystretch}{0.98}
\resizebox{0.58\textwidth}{!}{
\begin{tabular}{lcc}
\hline
\textbf{Type} & \textbf{Quantity} & \textbf{Source} \\
\hline
OXE & 3000 h & Open-source \\
AgiBot & 3276 h & Open-source \\
Droid & 350 h & Open-source \\
Robomind & 305 h & Open-source \\
Galaxea & 500 h & Open-source \\
Bridge V2 & 127 h & Open-source \\
HoloAssist & 166 h & Open-source \\
Ego4D & 3670 h & Open-source \\
EgoDex & 830 h & Open-source \\
HOI4D & 7.6 h & Open-source \\
Something-Something V2 & 240 h & Open-source \\
EgoVid & 556 h & Open-source \\
Egocentric-10k & 10000 h & Open-source \\
\hline
\end{tabular}
}
\end{table}

\paragraph{Discrete action supervision.}
To use human manipulation videos and web manipulation videos without robot
action annotations, we train a causal VQ-VAE to convert visual dynamics into
latent action IDs. The tokenizer is designed to encode action-induced changes
between consecutive frames rather than all appearance details. It first applies
a causal visual encoder to obtain temporally ordered visual features, then uses
latent queries and vector quantization to produce compact discrete codes. These
codes are used as discrete action supervision during Embodied Pretraining,
allowing the model to learn object interaction, action effects, and coarse
manipulation patterns from videos without robot action labels.

The tokenizer uses a codebook of size 16, organized as two groups with eight
entries in each group. For each consecutive-frame transition, it extracts four
latent action codes to represent the current visual change. Grouping the
codebook does not increase the number of codes per transition; instead, it
structures the code selection space. Compared with selecting codes directly from
a flat 16-entry codebook, this grouped design is easier to optimize and helps
alleviate codebook collapse and overfitting. During training, the decoder
reconstructs subsequent visual changes from the initial visual feature and the
quantized latent action codes, encouraging the codes to preserve action-relevant
dynamics such as object displacement, contact changes, hand or robot motion, and
scene-state transitions. After training, the decoder is discarded, and only the
causal encoder, attention module, and quantizer are kept for latent-action ID
extraction.

\begin{figure}[!htbp]
  \centering
  \includegraphics[width=0.72\linewidth]{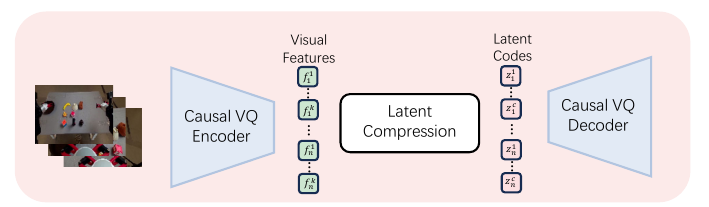}
  \vspace{-1mm}
  \caption{\small{Overview of the causal VQ-VAE latent action tokenizer. The
  tokenizer compresses action-induced visual changes into discrete latent action
  IDs, which are then used as supervision for Embodied Pretraining on videos
  without robot action labels. For \(f_i^j\), the subscript \(i\) denotes the
  frame ID and the superscript \(j\) denotes the image-feature patch ID. For
  \(z_i^j\), the subscript \(i\) denotes the frame ID and the superscript \(j\)
  denotes the codebook index.}}
  \label{fig:vqvae_overview}
  \vspace{-2mm}
\end{figure}

\begin{figure}[!htbp]
  \centering
  \includegraphics[width=0.30\linewidth]{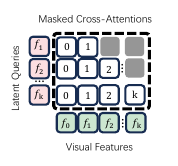}
  \vspace{-2mm}
\caption{\small{Masked cross-attention used in the causal VQ-VAE latent action
  tokenizer. Each query corresponds to an image frame, visual features are
  arranged in temporal order, and the query for a given frame can only attend to
  current and previous visual features. Gray cells denote future positions masked
  by the causal attention rule. The video frames are sampled at 5 Hz, while robot
  actions are represented at 30 Hz.}}
  \label{fig:vqvae_masked_cross_attention}
  \vspace{-4mm}
\end{figure}

The masked cross-attention mechanism in the causal VQ-VAE is used to prevent
future information leakage when extracting latent action IDs. As shown in
Figure~\ref{fig:vqvae_masked_cross_attention}, each query corresponds to an
image frame and can only attend to visual features from the current and previous
frames, while future-frame features are masked out. The robot action stream is
represented at 30 Hz, but the latent action tokenizer models consecutive video
frames sampled at 5 Hz. If the image stream were also sampled at 30 Hz,
autoregressive inference over a one-second horizon would produce up to
\(30\times4\) latent codes, which would substantially slow down decoding.
Moreover, overly dense image frames contain highly redundant visual information
and can weaken latent action learning. The 5 Hz frame rate therefore retains
short-horizon visual dynamics while keeping the latent code sequence compact.

\paragraph{Training configuration.}
The training procedure follows the Coarse-to-Fine Policy Learning design in the
main paper. Embodied Pretraining first trains the VLM to acquire embodied
grounding, task understanding, and discrete action semantics from VLM-related
data, manipulation videos, and robot trajectories. Continuous Policy Learning
then connects the learned action-semantic representation to the
gradient-decoupled DiT-based action expert using latent action tokens, while
retaining 40\% of the VLM-related data to reduce catastrophic forgetting.
Downstream fine-tuning uses task-specific demonstrations; for LIBERO, the
configured dataset is libero-union and the policy is trained with Relative
End-effector Action.

For training scale, Embodied Pretraining uses 240 H200 GPUs with data
parallelism and bf16 mixed precision, a per-GPU batch size of 8, and 350k
iterations. Continuous Policy Learning also uses 240 H200 GPUs with data
parallelism and bf16 mixed precision, a per-GPU batch size of 4, and 700k
iterations. Downstream fine-tuning is task-dependent; for LIBERO, we use 56 H200
GPUs with a per-GPU batch size of 4 and train for 12k iterations.

All training objectives share the same basic optimization and data-processing
settings. We use the AdamW optimizer with learning rate $2\times10^{-5}$, weight
decay 0, maximum gradient norm 1.0, a constant learning-rate schedule with 0.03
warmup ratio, bf16 mixed precision, gradient checkpointing, image augmentation,
random seed 42, and checkpoint interval 10k steps.

Manipulation data use image resolution 448 and the bimanual data format. For
single-arm robot data, the active arm is randomly mapped to the left or right
arm and padded into the bimanual format. The standard camera configuration
contains one head view and two wrist views. For robot datasets with non-standard
camera configurations, missing views are padded to this standard format, and the
padded view tokens are masked during attention computation by the
attention-level modality dropping mechanism. The action representation uses
Relative End-effector Action,
anchored-relative control, and rmat6d rotation in our experiments. Each
prediction covers a 1 s action chunk with 30 steps, corresponding to 30 Hz
action prediction. During training, we enable latent-action supervision, FAST
Token supervision, semantic-aligned 3D integration, and Embodiment
Canonicalization.
FAST Token supervision is used as an auxiliary training objective.

To improve robustness to missing or unreliable low-level inputs, we apply
attention-level modality dropping during training. The action-prediction queries
associated with FAST-style action tokens and the DiT-based action expert
cross-attend to the input images and robot-state tokens derived from
embodiment-aware state inputs. After the
cross-attention matrix is computed, we randomly drop selected attention weights:
with probability 0.2, we set the attention weights from the action-prediction
queries to the wrist-view image tokens to zero; with probability 0.2, we set the
attention weights to the robot-state tokens to zero; with probability
0.2, we drop both the wrist-view image tokens and robot-state tokens;
and with probability 0.4, we keep the full attention matrix unchanged.

\appendixsection{Embodied Reasoning Capability Analysis}
\label{sec:appendix_embodied_capability}

\paragraph{Embodied and spatial capability evaluation.}
Table~\ref{tab:embodied_qa} evaluates the embodied and spatial capabilities
learned before downstream continuous policy learning. In the Coarse-to-Fine
Policy Learning strategy, Embodied VLM Pre-training is designed as an easy-to-hard
curriculum for gradually improving the VLM's embodied reasoning and embodied
perception abilities before introducing low-level continuous actions. Embodied
VLM Pre-training uses all general vision-language, spatial perception, and
spatial understanding image-text data, together with trajectory reasoning,
pointing, mask tracking, and manipulation videos, under autoregressive
supervision. This
curriculum first preserves broad image-text understanding, then strengthens
spatial grounding and object-centric perception, and finally teaches the VLM
task-oriented planning, affordance understanding, and action-related visual
dynamics.

The results in Table~\ref{tab:embodied_qa} show that Embodied VLM Pre-training
improves embodied reasoning and embodied perception, reducing the gap between a
general VLM and the action semantics required for directly solving VLA tasks,
thereby satisfying the coarse-to-fine training requirement before continuous
action learning. The learned representation improves task-oriented planning
and affordance
understanding, as well as spatial grounding and trajectory reasoning; meanwhile,
the results on general VLM benchmarks such as CVBench show that the original
general image-text understanding ability of the VLM is preserved. This balance
allows the model to keep the semantic strengths of the VLM while making the
representation more suitable for downstream VLA policy learning.

\begin{table}[!htbp]
\centering
\caption{General QA and embodied capability evaluation.}
\label{tab:embodied_qa}
\scriptsize
\setlength{\tabcolsep}{3pt}
\resizebox{\textwidth}{!}{
\begin{tabular}{lcccccccccc}
\hline
\thead{Method} & \thead{EgoPlan2} & \thead{Where2Place} & \thead{RefSpatial-\\bench} & \thead{ShareRobot-\\affordance} & \thead{ShareRobot-\\trajectory} & \thead{BLINK} & \thead{EmbSpatial} & \thead{ERQA} & \thead{SAT} & \thead{CVBench} \\
\hline
Embodied-R1-7B & 30.82 & 69.5 & 39.5 & 22.69 & 63.29 & 66.73 & 67.4 & 39.1 & 70.0 & 82.7 \\
Qwen3-VL-8B & 39.06 & 21.2 & 35.75 & 43.72 & 71.21 & 84.08 & 78.9 & \textbf{43.5} & 53.33 & \textbf{86.32} \\
InternVLA-M1 & 30.58 & 52.19 & 35.5 & 18.99 & 42.33 & 65.65 & 65.27 & 40.35 & 68.0 & 74.41 \\
InternVL3-8B & 41.03 & 12.0 & 18.0 & 14.45 & 47.0 & 61.36 & 63.63 & 32.08 & 52.67 & 75.36 \\
RoboBrain2.0-7B & 33.23 & 63.59 & 32.5 & 28.05 & 44.88 & 83.95 & 76.32 & 34.25 & \textbf{75.33} & 85.75 \\
\rowcolor{oursgray}\textbf{Ours-8B} & \textbf{47.00} & \textbf{70.23} & \textbf{51.5} & \textbf{59.61} & \textbf{81.74} & \textbf{85.69} & \textbf{78.92} & \textbf{43.5} & 72.0 & 85.33 \\
\hline
\end{tabular}
}
\end{table}

The curriculum-order ablation in Table~\ref{tab:curriculum_order_ablation}
further shows that deviating from the easy-to-hard SU \(\rightarrow\) SP order
causes capability forgetting: the reversed SP \(\rightarrow\) SU order lowers
BLINK and degrades grounded perception metrics such as Where2Place, Refloc, and
Affordance.

\begin{table}[!htbp]
\centering
\caption{\textbf{Curriculum-learning order ablation.} SU denotes spatial understanding data, and SP denotes spatial perception data. The SU \(\rightarrow\) SP curriculum reports the observed range across validated standard-order configurations. Reversing the order causes capability forgetting on spatial reasoning and grounded perception benchmarks.}
\label{tab:curriculum_order_ablation}
\scriptsize
\setlength{\tabcolsep}{5pt}
\renewcommand{\arraystretch}{1.08}
\resizebox{0.78\textwidth}{!}{
\begin{tabular}{lccccc}
\hline
\textbf{Training Order} & \textbf{BLINK} & \textbf{SAT} & \textbf{Where2Place} & \textbf{Refloc} & \textbf{Affordance} \\
\hline
SU \(\rightarrow\) SP (Ours Curriculum) & 82--85 & 64--72 & 60--70 & 55--62 & 37--59 \\
SP \(\rightarrow\) SU (Reversed Curriculum) & 78.71 & 70.67 & 57.07 & 35.00 & 29.60 \\
\hline
\end{tabular}
}
\end{table}

\appendixsection{Simulation Benchmark Results}
\label{sec:appendix_additional_simulation}

\paragraph{Simulation benchmarks.}
Tables~\ref{tab:libero}, \ref{tab:libero_plus}, \ref{tab:robotwin}, and
\ref{tab:robotwin_part2} provide the expanded simulation results behind the
aggregate comparison in the main paper. The RoboTwin 2.0 benchmark is split into
two tables for readability: Part I reports the first group of tasks, and Part II
reports the remaining tasks together with the clean and randomized averages.

\begin{table}[!htbp]
\centering
\caption{LIBERO benchmark evaluation.}
\label{tab:libero}
\scriptsize
\setlength{\tabcolsep}{4pt}
\begin{tabular}{lccccc}
\hline
\thead{Method} & \thead{Spatial} & \thead{Object} & \thead{Goal} & \thead{Long} & \thead{Average} \\
\hline
Octo & 78.9 & 85.7 & 84.6 & 51.1 & 75.1 \\
Diffusion Policy & 78.5 & 87.5 & 73.5 & 64.8 & 76.1 \\
OpenVLA & 84.7 & 88.4 & 79.2 & 53.7 & 76.5 \\
SpatialVLA & 88.2 & 89.9 & 78.6 & 55.5 & 78.1 \\
WorldVLA (512x512) & 87.6 & 96.2 & 83.4 & 60.0 & 81.8 \\
CoT-VLA & 87.5 & 91.6 & 87.6 & 69.0 & 83.9 \\
DreamVLA & 97.5 & 94.0 & 89.5 & 89.5 & 92.6 \\
$\pi_0$ & 98.0 & 96.8 & 94.4 & 88.4 & 94.4 \\
UniVLA & 95.4 & 98.8 & 93.6 & 94.0 & 95.5 \\
Discrete Diffusion VLA & 97.2 & 98.6 & 97.4 & 92.0 & 96.3 \\
$\pi_{0.5}$ & 98.8 & 98.2 & 98.0 & 92.4 & 96.9 \\
GR00T-N1.6 & 97.7 & 98.5 & 97.5 & 94.4 & 97.0 \\
OpenVLA-OFT & 97.6 & 98.4 & 97.9 & 94.5 & 97.1 \\
X-VLA & 98.2 & 98.6 & 97.8 & \textbf{97.6} & 98.1 \\
ACoT & 99.4 & 99.6 & \textbf{98.8} & 96.0 & 98.5 \\
\rowcolor{oursgray}\textbf{Ours} & \textbf{99.7} & \textbf{99.8} & 98.4 & 96.8 & \textbf{98.7} \\
\hline
\end{tabular}
\end{table}

\begin{table}[!htbp]
\centering
\caption{LIBERO-Plus benchmark evaluation.}
\label{tab:libero_plus}
\scriptsize
\setlength{\tabcolsep}{4pt}
\begin{tabular}{lcccccccc}
\hline
\thead{Method} & \thead{Camera} & \thead{Robot} & \thead{Language} & \thead{Light} & \thead{Background} & \thead{Noise} & \thead{Layout} & \thead{Average} \\
\hline
WorldVLA & 0.1 & 27.9 & 41.6 & 43.7 & 17.1 & 10.9 & 38.0 & 25.0 \\
OpenVLA & 0.8 & 3.5 & 23.0 & 8.1 & 34.8 & 15.2 & 28.5 & 15.6 \\
OpenVLA-OFT & 56.4 & 31.9 & 79.5 & 88.7 & 93.3 & 75.8 & 74.2 & 69.6 \\
UniVLA & 1.8 & 46.2 & 69.6 & 69.0 & 81.0 & 21.2 & 31.9 & 42.9 \\
NORA & 2.2 & 37.0 & 65.1 & 45.7 & 58.6 & 12.8 & 62.1 & 39.0 \\
$\pi_0$-Fast & 65.1 & 21.6 & 61.0 & 73.2 & 73.2 & 74.4 & 68.8 & 61.6 \\
$\pi_0$ & 61.0 & 40.8 & 63.5 & 89.3 & 84.1 & 80.1 & 76.4 & 69.4 \\
$\pi_{0.5}$ & 75.8 & 79.4 & 83.3 & 95.5 & 95.0 & 89.6 & 87.0 & 85.7 \\
ACoT & 72.6 & 82.6 & \textbf{87.5} & 97.7 & \textbf{96.5} & 87.8 & 88.1 & 86.6 \\
\rowcolor{oursgray}\textbf{Ours} & \textbf{82.6} & \textbf{84.1} & 82.4 & \textbf{97.9} & 93.1 & \textbf{90.0} & \textbf{89.4} & \textbf{88.7} \\
\hline
\end{tabular}
\end{table}

\begin{table}[!htbp]
\centering
\caption{\textbf{RoboTwin 2.0 benchmark evaluation (Part I).} Clean/Rand denote clean/randomized settings; the full benchmark is split into two tables for readability.}
\label{tab:robotwin}
\scriptsize
\setlength{\tabcolsep}{3pt}
\newcommand{\robotwinGroupRuleLength}{1.05cm}
\newcommand{\robotwinGroupHead}[1]{\raisebox{-2.5ex}{\begin{tabular}[c]{@{}c@{}}#1\\[-0.25ex]\rule{\robotwinGroupRuleLength}{0.35pt}\end{tabular}}}
\begin{tabular}{lcccccc>{\columncolor{oursgray}}c>{\columncolor{oursgray}}c}
\hline
\multirow{2}{*}{\thead{Task}} & \multicolumn{2}{c}{\robotwinGroupHead{$\pi_{0.5}$}} & \multicolumn{2}{c}{\robotwinGroupHead{X-VLA}} & \multicolumn{2}{c}{\robotwinGroupHead{ACoT}} & \multicolumn{2}{>{\columncolor{oursgray}}c}{\robotwinGroupHead{\textbf{Ours}}} \\
 & \thead{Clean} & \thead{Rand} & \thead{Clean} & \thead{Rand} & \thead{Clean} & \thead{Rand} & \thead{\textbf{Clean}} & \thead{\textbf{Rand}} \\
\hline
Adjust Bottle & 100 & 99 & 100 & 99 & 99 & 99 & 90 & 92 \\
Beat Block Hammer & 96 & 93 & 92 & 88 & 82 & 85 & 90 & 87 \\
Blocks Ranking RGB & 92 & 85 & 83 & 83 & 86 & 83 & 97 & 96 \\
Blocks Ranking Size & 49 & 26 & 67 & 74 & 56 & 62 & 88 & 82 \\
Click Alarmclock & 98 & 89 & 99 & 99 & 97 & 96 & 96 & 98 \\
Click Bell & 99 & 66 & 100 & 100 & 98 & 99 & 100 & 99 \\
Dump Bin Bigbin & 92 & 97 & 79 & 77 & 90 & 86 & 89 & 92 \\
Grab Roller & 100 & 100 & 100 & 100 & 100 & 100 & 100 & 100 \\
Handover Block & 66 & 57 & 73 & 37 & 68 & 31 & 94 & 82 \\
Handover Mic & 98 & 97 & 0 & 0 & 94 & 90 & 85 & 79 \\
Hanging Mug & 18 & 17 & 23 & 27 & 14 & 12 & 47 & 38 \\
Lift Pot & 96 & 85 & 99 & 100 & 87 & 84 & 97 & 99 \\
Move Can Pot & 51 & 55 & 89 & 86 & 76 & 80 & 90 & 94 \\
Move Pillbottle Pad & 84 & 61 & 73 & 71 & 78 & 78 & 96 & 98 \\
Move Playingcard Away & 96 & 84 & 93 & 98 & 88 & 78 & 100 & 98 \\
Move Stapler Pad & 56 & 42 & 78 & 73 & 76 & 68 & 89 & 84 \\
Open Laptop & 90 & 96 & 93 & 100 & 88 & 89 & 92 & 93 \\
Open Microwave & 34 & 77 & 79 & 71 & 74 & 78 & 86 & 87 \\
Pick Diverse Bottles & 81 & 71 & 58 & 36 & 58 & 63 & 86 & 88 \\
Pick Dual Bottles & 93 & 63 & 47 & 36 & 55 & 69 & 97 & 93 \\
Place A2B Left & 87 & 82 & 48 & 49 & 80 & 78 & 88 & 89 \\
Place A2B Right & 87 & 84 & 36 & 36 & 82 & 77 & 88 & 86 \\
Place Bread Basket & 77 & 64 & 81 & 71 & 73 & 82 & 94 & 95 \\
Place Bread Skillet & 85 & 66 & 77 & 67 & 78 & 78 & 93 & 97 \\
Place Burger Fries & 94 & 87 & 94 & 94 & 93 & 89 & 96 & 97 \\
\hline
\end{tabular}
\end{table}

\begin{table}[!htbp]
\centering
\caption{\textbf{RoboTwin 2.0 benchmark evaluation (Part II).} Clean/Rand denote clean/randomized settings; bold marks the best Average within each setting.}
\label{tab:robotwin_part2}
\scriptsize
\setlength{\tabcolsep}{3pt}
\newcommand{\robotwinPartTwoGroupRuleLength}{1.05cm}
\newcommand{\robotwinPartTwoGroupHead}[1]{\raisebox{-2.5ex}{\begin{tabular}[c]{@{}c@{}}#1\\[-0.25ex]\rule{\robotwinPartTwoGroupRuleLength}{0.35pt}\end{tabular}}}
\begin{tabular}{lcccccc>{\columncolor{oursgray}}c>{\columncolor{oursgray}}c}
\hline
\multirow{2}{*}{\thead{Task}} & \multicolumn{2}{c}{\robotwinPartTwoGroupHead{$\pi_{0.5}$}} & \multicolumn{2}{c}{\robotwinPartTwoGroupHead{X-VLA}} & \multicolumn{2}{c}{\robotwinPartTwoGroupHead{ACoT}} & \multicolumn{2}{>{\columncolor{oursgray}}c}{\robotwinPartTwoGroupHead{\textbf{Ours}}} \\
 & \thead{Clean} & \thead{Rand} & \thead{Clean} & \thead{Rand} & \thead{Clean} & \thead{Rand} & \thead{\textbf{Clean}} & \thead{\textbf{Rand}} \\
\hline
Place Can Basket & 62 & 62 & 49 & 52 & 68 & 66 & 85 & 84 \\
Place Cans Plasticbox & 94 & 84 & 97 & 98 & 96 & 98 & 99 & 96 \\
Place Container Plate & 99 & 95 & 97 & 95 & 95 & 100 & 98 & 99 \\
Place Dual Shoes & 75 & 75 & 79 & 88 & 78 & 82 & 93 & 87 \\
Place Empty Cup & 100 & 99 & 100 & 98 & 96 & 97 & 99 & 99 \\
Place Fan & 87 & 85 & 80 & 75 & 77 & 85 & 95 & 91 \\
Place Mouse Pad & 60 & 39 & 70 & 70 & 60 & 64 & 84 & 85 \\
Place Object Basket & 80 & 76 & 44 & 39 & 72 & 75 & 87 & 92 \\
Place Object Scale & 86 & 80 & 52 & 74 & 82 & 80 & 92 & 89 \\
Place Object Stand & 91 & 85 & 86 & 88 & 88 & 90 & 98 & 95 \\
Place Phone Stand & 81 & 81 & 88 & 87 & 78 & 78 & 93 & 91 \\
Place Shoe & 92 & 93 & 96 & 95 & 93 & 91 & 97 & 98 \\
Press Stapler & 87 & 83 & 92 & 98 & 90 & 87 & 83 & 80 \\
Put Bottles Dustbin & 84 & 79 & 74 & 77 & 77 & 71 & 84 & 88 \\
Put Object Cabinet & 80 & 79 & 46 & 48 & 76 & 60 & 86 & 78 \\
Rotate QRcode & 89 & 87 & 34 & 33 & 78 & 60 & 92 & 84 \\
Scan Object & 72 & 65 & 14 & 36 & 48 & 50 & 83 & 79 \\
Shake Bottle Horizontally & 99 & 99 & 100 & 100 & 100 & 97 & 98 & 95 \\
Shake Bottle & 99 & 97 & 99 & 100 & 98 & 99 & 97 & 97 \\
Stack Blocks Three & 91 & 76 & 6 & 10 & 68 & 74 & 94 & 96 \\
Stack Blocks Two & 97 & 100 & 92 & 87 & 92 & 89 & 97 & 96 \\
Stack Bowls Three & 77 & 71 & 76 & 86 & 84 & 72 & 85 & 88 \\
Stack Bowls Two & 95 & 96 & 96 & 93 & 97 & 97 & 95 & 98 \\
Stamp Seal & 79 & 55 & 76 & 82 & 78 & 82 & 94 & 95 \\
Turn Switch & 62 & 54 & 40 & 61 & 64 & 58 & 77 & 73 \\
Average & 82.74 & 76.76 & 72.88 & 72.84 & 80.06 & 78.72 & \textbf{91.06} & \textbf{89.92} \\
\hline
\end{tabular}
\end{table}

\appendixsection{Ablation Experiments}
\label{sec:appendix_detailed_ablation}

\paragraph{Overview.}
This section provides supplementary ablations that are not repeated in the main
paper: attention-level modality dropping, Embodiment Canonicalization,
latent action modeling, and a feature visualization for semantic-aligned 3D
integration.

\paragraph{Attention-level modality dropping.}
Table~\ref{tab:ablation_modality_dropout} evaluates the wrist-view and
robot-state dropping strategy used during training. Without dropping,
the model can over-rely on low-level wrist observations or robot states,
which weakens the generality of the learned representation. Dropping wrist-view
attention encourages the model to use head-view global context, improving
layout robustness. Dropping state attention makes the model rely more on visual
evidence, which improves robustness under robot, camera, lighting, background,
and noise variations. Combining both drops yields the best average performance.

\begin{table}[!htbp]
\centering
\caption{\textbf{Attention-level modality dropping ablation.} All entries are success rates (\%) on LIBERO-Plus. Wrist denotes wrist-view image tokens, and State denotes proprioceptive state tokens.}
\label{tab:ablation_modality_dropout}
\scriptsize
\setlength{\tabcolsep}{4pt}
\renewcommand{\arraystretch}{1.08}
\resizebox{0.90\textwidth}{!}{
\begin{tabular}{lcccccccc}
\hline
\textbf{Method} & \textbf{Cam.} & \textbf{Rob.} & \textbf{Lang.} & \textbf{Light} & \textbf{Bg.} & \textbf{Noise} & \textbf{Layout} & \textbf{Avg.} \\
\hline
w/o Dropping & 81.4 & 82.9 & \textbf{82.6} & 96.8 & 92.6 & 88.7 & 88.1 & 87.7 \\
Only Drop Wrist & 81.2 & 83.1 & 82.4 & 97.2 & 92.9 & 89.2 & \textbf{89.5} & 88.1 \\
Only Drop State & 82.4 & \textbf{84.2} & 82.2 & \textbf{97.9} & \textbf{93.7} & 89.8 & 88.4 & 88.5 \\
\rowcolor{oursgray}\textbf{Ours (Drop Wrist + Drop State)} & \textbf{82.6} & 84.1 & 82.4 & \textbf{97.9} & 93.1 & \textbf{90.0} & 89.4 & \textbf{88.7} \\
\hline
\end{tabular}
}
\end{table}

\paragraph{Embodiment Canonicalization.}
Table~\ref{tab:ablation_state_aware} studies how embodiment information should
be injected. Removing all embodiment-specific settings weakens performance, but
adding X-VLA-style soft prompts hurts more because it can entangle robot
identity with the shared VLA representation under imbalanced robot data.
Embodiment-aware state adaptation is more effective because robot-specific
variation is adapted through the low-level state interface, rather than being
injected as robot-specific semantic prompts into the shared VLA representation.
Output-aware conditioning does not bring additional benefit in this setting, and
the final design uses Embodiment Canonicalization without X-VLA-style soft
prompts.

\begin{table}[!htbp]
\centering
\caption{\textbf{Embodiment Canonicalization ablation.} All entries are success rates (\%) on LIBERO-Plus.}
\label{tab:ablation_state_aware}
\scriptsize
\setlength{\tabcolsep}{4pt}
\renewcommand{\arraystretch}{1.08}
\resizebox{\textwidth}{!}{
\begin{tabular}{lcccccccc}
\hline
\textbf{Method} & \textbf{Cam.} & \textbf{Rob.} & \textbf{Lang.} & \textbf{Light} & \textbf{Bg.} & \textbf{Noise} & \textbf{Layout} & \textbf{Avg.} \\
\hline
\makecell[l]{No Embodiment Canonicalization\\(w/o all embodiment-specific modules)} & 80.3 & 82.7 & 80.1 & 95.0 & 91.7 & 89.1 & 88.1 & \mbox{86.9 {\color{red}{(-1.8)}}} \\
\makecell[l]{w/ X-VLA Soft Prompt\\(Soft Prompt + Embodiment-Aware State + Output-Aware)} & 78.2 & 79.4 & 78.9 & 93.9 & 90.4 & 86.9 & 86.1 & \mbox{85.0 {\color{red}{(-3.7)}}} \\
\makecell[l]{Embodiment-Aware State and Output-Aware\\(w/o Soft Prompt)} & 81.5 & 83.8 & 81.8 & 97.3 & 92.9 & 89.5 & 88.9 & \mbox{88.1 {\color{red}{(-0.6)}}} \\
\makecell[l]{w/o Embodiment-Specific State Projector\\(Output-Aware only)} & 80.4 & 82.4 & 79.9 & 95.4 & 91.3 & 88.8 & 87.5 & \mbox{86.7 {\color{red}{(-2.0)}}} \\
\rowcolor{oursgray}\textbf{Ours (Embodiment-Aware State)} & \textbf{82.6} & \textbf{84.1} & \textbf{82.4} & \textbf{97.9} & \textbf{93.1} & \textbf{90.0} & \textbf{89.4} & \textbf{88.7} \\
\hline
\end{tabular}
}
\end{table}

\paragraph{Latent action modeling.}
Table~\ref{tab:ablation_lam} compares latent action supervision variants.
Removing latent action IDs reduces the diversity and continuity of action
semantics learned from action-free videos. Encoding only the initial and final
states, as in LAPA-style supervision, provides coarser action information. Our
continuous latent action modeling uses five frames within one second and better
captures short-horizon action dynamics, leading to the best performance.

\begin{table}[!htbp]
\centering
\caption{\textbf{Latent action modeling ablation.} All entries are success rates (\%) on LIBERO-Plus.}
\label{tab:ablation_lam}
\scriptsize
\setlength{\tabcolsep}{4pt}
\renewcommand{\arraystretch}{1.08}
\resizebox{\textwidth}{!}{
\begin{tabular}{lcccccccc}
\hline
\textbf{Method} & \textbf{Cam.} & \textbf{Rob.} & \textbf{Lang.} & \textbf{Light} & \textbf{Bg.} & \textbf{Noise} & \textbf{Layout} & \textbf{Avg.} \\
\hline
w/o Latent Action IDs & 81.2 & 82.4 & 79.7 & 95.8 & 91.8 & 87.9 & 88.9 & 87.1 {\color{red}{(-1.6)}} \\
\makecell[l]{LAPA-style Latent Actions\\(initial/final states only)} & 82.3 & 83.6 & 81.2 & 96.9 & 92.7 & 89.7 & 88.9 & 88.1 {\color{red}{(-0.6)}} \\
\rowcolor{oursgray}\textbf{\makecell[l]{Ours (Continuous Latent Actions\\with 5 Frames in 1 s)}} & \textbf{82.6} & \textbf{84.1} & \textbf{82.4} & \textbf{97.9} & \textbf{93.1} & \textbf{90.0} & \textbf{89.4} & \textbf{88.7} \\
\hline
\end{tabular}
}
\end{table}

\begin{figure}[!htbp]
  \centering
  \includegraphics[width=0.75\linewidth]{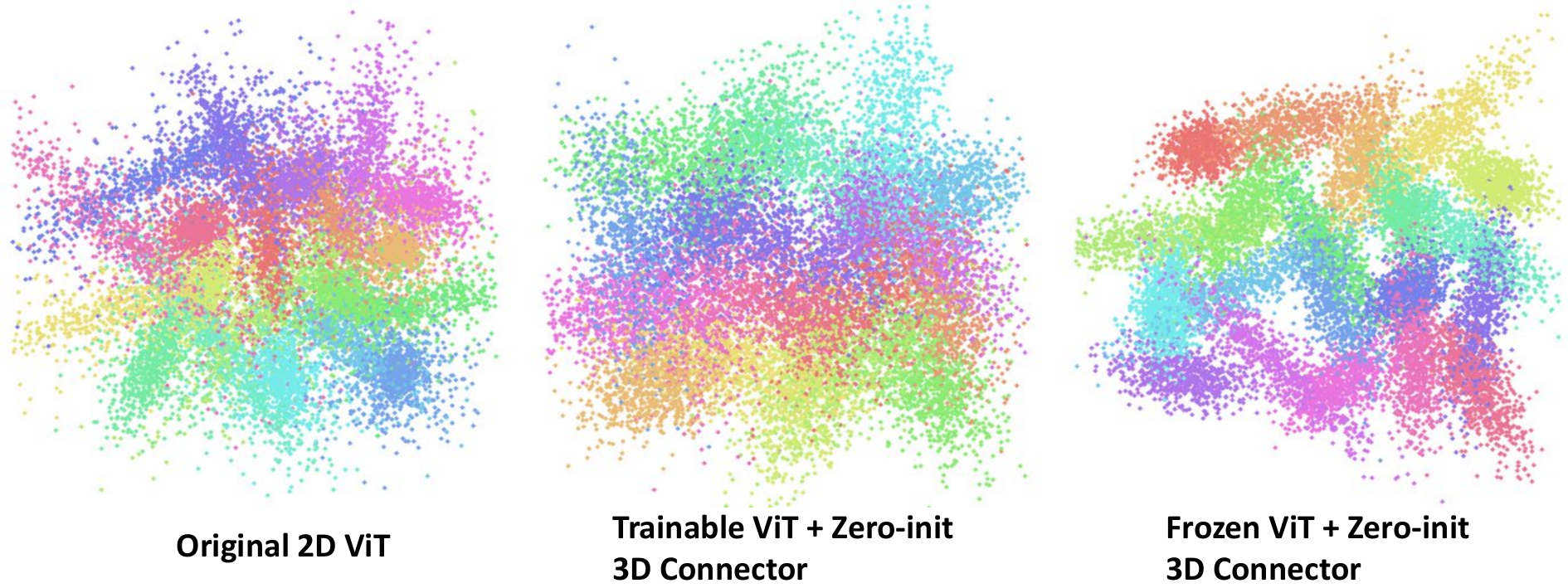}
  \caption{\small{Feature visualization for the VGGT ablation on 20 ImageNet
  classes, with 1,000 samples per class. (a) Original ViT. (b) Unfrozen ViT +
  Zero-init VGGT Connector. (c) Ours: Frozen ViT + Zero-init VGGT Connector.}}
  \label{fig:vggt_ablation}
\end{figure}

\paragraph{Geometry-aware visual encoding.}
Figure~\ref{fig:vggt_ablation} compares feature distributions for three visual
encoding variants on 20 ImageNet classes, with 1,000 samples per class.
Figure~\ref{fig:vggt_ablation}(a) shows that the Original ViT provides partial
semantic separation but still exhibits noticeable class overlap.
Figure~\ref{fig:vggt_ablation}(b) shows that unfreezing the original ViT while
adding a zero-initialized VGGT connector disrupts the pretrained semantic
feature space, leading to less stable class boundaries. In contrast,
Figure~\ref{fig:vggt_ablation}(c) freezes the original ViT and injects VGGT
features through the zero-initialized connector, yielding clearer inter-class
separation and more structured intra-class distributions. These results support
our semantic-aligned 3D integration design.

\appendixsection{Dual-Arm Data Efficiency}
\label{sec:appendix_real_world_dual_arm}

\paragraph{Data-efficiency evaluation.}
Table~\ref{tab:dual_arm_data_efficiency} evaluates real-world data efficiency
on AgileX by varying the amount of downstream training data from 25\% to 100\%.
The results demonstrate strong data efficiency. Ours achieves the best average
success rate under every data ratio, with especially clear gains in the
low-data regimes. This indicates that the learned geometry-aware action
representation remains effective under limited demonstrations on three
contact-rich bimanual manipulation tasks.

\begin{table}[!htbp]
\centering
\caption{\textbf{Dual-arm data-efficiency analysis on AgileX.} All entries are success rates (\%); bold indicates the best performance under each data ratio, and Average is weighted by task trials.}
\label{tab:dual_arm_data_efficiency}
\scriptsize
\renewcommand{\arraystretch}{1.06}
\setlength{\tabcolsep}{4pt}
\begin{tabular}{llccccc}
\toprule
\multirow{2}{*}{\textbf{Data Ratio}} 
& \multirow{2}{*}{\textbf{Task}} 
& \multicolumn{5}{c}{\textbf{Method}} \\
\cmidrule(lr){3-7}
& & $\boldsymbol{\pi_0}$ & \textbf{X-VLA} & $\boldsymbol{\pi_{0.5}}$ & \textbf{ACoT} & \textbf{Ours} \\
\midrule
\multirow{4}{*}{25\%}
& T-shirt Folding & 28.7 & 56.7 & 54.7 & 60.7 & \textbf{66.7} \\
& Shorts Folding  & 36.7 & 62.7 & 58.7 & 62.7 & \textbf{72.7} \\
& Parcel Flipping & 34.0 & 58.0 & 54.0 & 56.0 & \textbf{68.0} \\
& Average         & 33.1 & 59.1 & 55.8 & 59.8 & \textbf{69.1} \\
\midrule
\multirow{4}{*}{50\%}
& T-shirt Folding & 44.7 & 68.7 & 68.7 & 64.0 & \textbf{74.7} \\
& Shorts Folding  & 52.7 & 74.7 & 72.7 & 66.7 & \textbf{78.7} \\
& Parcel Flipping & 52.0 & 70.0 & 68.0 & 62.0 & \textbf{76.0} \\
& Average         & 49.8 & 71.1 & 69.8 & 64.2 & \textbf{76.5} \\
\midrule
\multirow{4}{*}{75\%}
& T-shirt Folding & 50.7 & 74.0 & 74.0 & 70.7 & \textbf{78.7} \\
& Shorts Folding  & 62.0 & 80.7 & 78.0 & 72.7 & \textbf{82.7} \\
& Parcel Flipping & 58.0 & 76.0 & 72.0 & 70.0 & \textbf{80.0} \\
& Average         & 56.9 & 76.9 & 74.7 & 71.1 & \textbf{80.5} \\
\midrule
\multirow{4}{*}{100\%}
& T-shirt Folding & 58.7 & 80.7 & 80.7 & 76.7 & \textbf{82.7} \\
& Shorts Folding  & 68.0 & 86.0 & 84.0 & 80.7 & \textbf{88.0} \\
& Parcel Flipping & 66.0 & 83.0 & 77.0 & 76.0 & \textbf{87.0} \\
& Average         & 64.2 & 83.2 & 80.6 & 77.8 & \textbf{85.9} \\
\bottomrule
\end{tabular}
\end{table}

\appendixsection{Universal Object Grasping Evaluation}
\label{sec:appendix_universal_grasping}

\paragraph{Mask-guided universal grasping.}
The real-world evaluation constructs a large-scale universal grasping benchmark
with unseen objects and cluttered scenes. The benchmark uses a referring
mask-guided prompt to specify arbitrary objects and tests whether the model can
correctly manipulate the object based on its spatial location and geometry. This
capability is supported by the mask-tracking task used during Embodied
Pretraining. In this task, an object is marked by a mask in the image at time
\(t_0\), and the model is trained to localize the same marked object in the
image at time \(t_1\). Therefore, the model cannot rely only on a static image
location, and must instead learn to associate the mask-indicated object instance
with its later appearance.

During universal object grasping, the system first takes the image at time
\(t_0\) and the user instruction, e.g., ``put the apple into the plate'', as
input to the VLM. Using the grounding capability learned during Embodied
Pretraining, the model predicts a bounding box for the object referred to by the
instruction. We then apply SAM to this grounded box to obtain an instance mask
for the target object. The original \(t_0\) image is converted into a grayscale
three-channel image, and the target mask is overlaid with a randomly sampled
color at 50\% transparency, forming a mask-guided image \(t_0'\). The language
instruction is rewritten to refer to the same color, e.g., ``put the object with
the \(x\)-colored mask into the plate''. The pair of \(t_0'\) and the rewritten
instruction, together with the current image at time \(t\), is then fed to the
same VLM, and the DiT-based action expert predicts the robot action. In this
process, the colored mask serves as a visual referring cue rather than an object
appearance cue, while the pretrained mask-tracking ability helps the model keep
attending to the specified object in subsequent observations.

\begin{figure}[!htbp]
  \centering
  \includegraphics[width=0.78\linewidth]{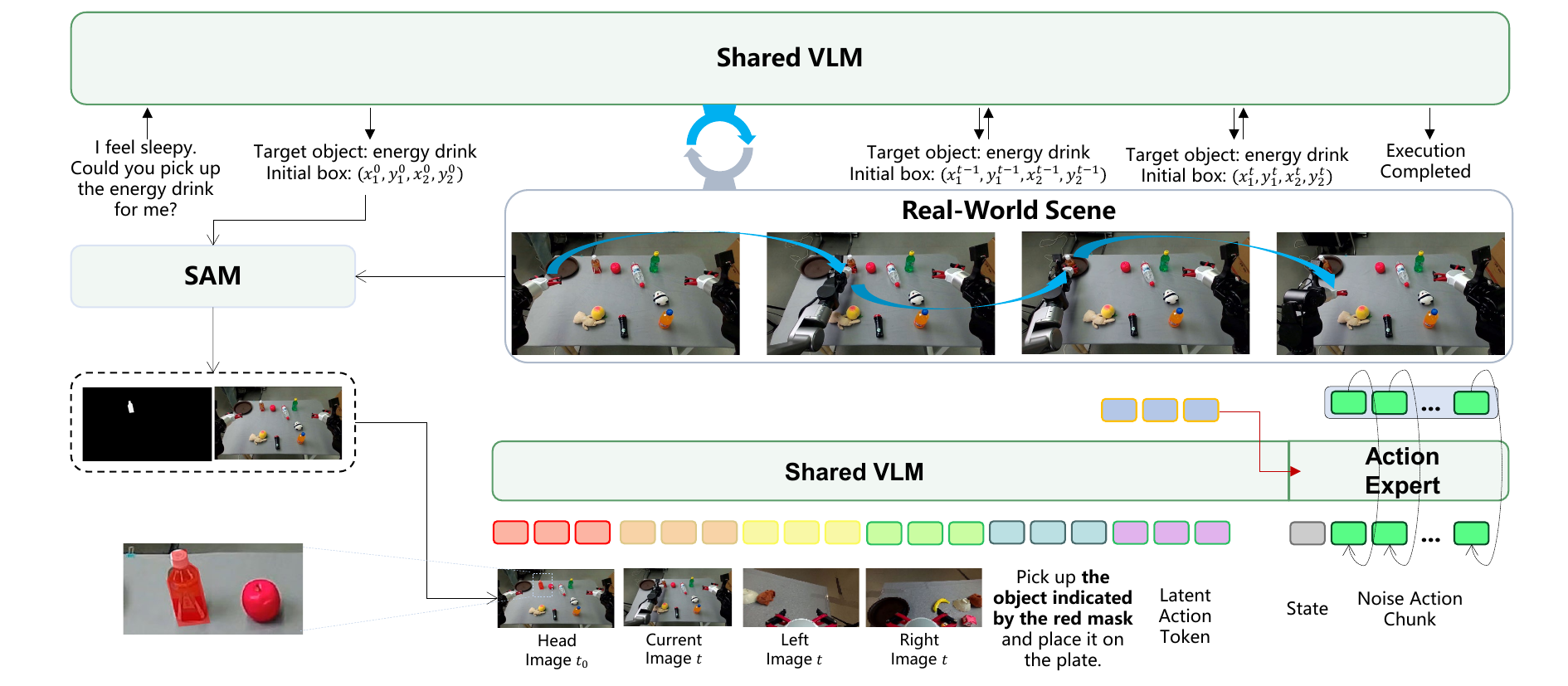}
  \vspace{-1mm}
  \caption{\small{Mask-guided universal object grasping pipeline. The same VLM
  is used for target grounding and mask-guided reasoning. Given the first-frame
  target mask and current multi-view observation, the DiT-based action expert
  predicts the robot action for manipulating the specified object.}}
  \label{fig:universal_grasping_pipeline}
  \vspace{-3mm}
\end{figure}

\paragraph{Object split.}
Table~\ref{tab:universal_grasping_object_split} summarizes the object split used
in this benchmark. The training split contains 35 target objects, with 100
demonstrations collected for each object. During data collection, each scene
contains the target object and 4--20 background objects sampled from the
training-object pool, including objects with similar object categories. The test
split contains 212 target objects that do not overlap with the training objects,
covering the same object categories to evaluate generalization over object
geometry rather than memorization of specific object instances.

\begin{table}[!htbp]
\centering
\caption{\textbf{Object split for the real-world universal object grasping benchmark.} The training split contains 35 target objects with 100 demonstrations per object, and the test split contains 212 unseen target objects; object categories follow the main-paper taxonomy and evaluate geometry generalization rather than object memorization.}
\label{tab:universal_grasping_object_split}
\scriptsize
\setlength{\tabcolsep}{6pt}
\renewcommand{\arraystretch}{1.00}
\vspace{-1mm}

\begin{minipage}[t]{0.49\textwidth}
\centering
\textbf{Training Objects}\\[-0.5mm]
\resizebox{\linewidth}{!}{
\begin{tabular}{lc}
\hline
\multicolumn{1}{l}{\thead{Object Category}} & \thead{Number of Objects} \\
\hline
Axisymmetric & 16 \\
Block-like & 6 \\
Irregular & 10 \\
Tool & 1 \\
Bagged & 2 \\
\textbf{Total} & \textbf{35} \\
\hline
\end{tabular}
}
\end{minipage}
\hfill
\begin{minipage}[t]{0.49\textwidth}
\centering
\textbf{Test Objects}\\[-0.5mm]
\resizebox{\linewidth}{!}{
\begin{tabular}{lc}
\hline
\multicolumn{1}{l}{\thead{Object Category}} & \thead{Number of Objects} \\
\hline
Axisymmetric & 104 \\
Block-like & 54 \\
Irregular & 43 \\
Tool & 6 \\
Bagged & 5 \\
\textbf{Total} & \textbf{212} \\
\hline
\end{tabular}
}
\end{minipage}

\end{table}

\paragraph{Evaluation protocol.}
Each test object is evaluated for 10 trials. In every trial, we randomize the
target position, target orientation, number of background objects, background
object categories, background object positions, and tabletop layout. This gives
2,120 real-robot trials per method in a single evaluation setting. We evaluate
three settings: sparse clutter with 4--8 background objects, dense clutter with
8--20 background objects, and background/lighting variation using the same dense
layout protocol. In the background/lighting variation setting, the tablecloth
color and texture are changed every 100 trials, and lighting attributes such as
color, brightness, temperature, and direction are changed every 10 trials. Across
the three settings, each method is evaluated over 6,360 real-robot trials.

\end{document}